\documentclass{article}

     \usepackage[final, nonatbib]{tackling_climate_workshop_style}

\usepackage[utf8]{inputenc} 
\usepackage[T1]{fontenc}    
\usepackage{hyperref}       
\usepackage{url}            
\usepackage{booktabs}       
\usepackage{amsfonts}       
\usepackage{nicefrac}       
\usepackage{microtype}      
\usepackage{graphicx}
\usepackage{multirow}
\usepackage{amsmath}
\usepackage{adjustbox}

\title{Zero-shot Microclimate Prediction with Deep Learning}

\author{
  Iman Deznabi \\
  Manning College of Information and Computer Sciences\\
  University of Massachusetts Amherst\\
  Amherst, MA 01002 \\
  \texttt{iman@cs.umass.edu} \\
  \And
  Peeyush Kumar \\
  Microsoft Research\\
  Redmond, WA 98052 \\
  \texttt{iman@cs.umass.edu} \\
  \And
  Madalina Fiterau \\
  Manning College of Information and Computer Sciences\\
  University of Massachusetts Amherst\\
  Amherst, MA 01002 \\
  \texttt{mfiterau@cs.umass.edu} \\
}

\begin{document}

\maketitle

\begin{abstract}
    Weather station data is a valuable resource for climate prediction, however, its reliability can be limited in remote locations. To compound the issue, making local predictions often relies on sensor data that may not be accessible for a new, previously unmonitored location. In response to these challenges, we propose a novel zero-shot learning approach designed to forecast various climate measurements at new and unmonitored locations. Our method surpasses conventional weather forecasting techniques in predicting microclimate variables by leveraging knowledge extracted from other geographic locations.
\end{abstract}

\section{Introduction}
Micro-climate prediction involves the forecasting and analysis of localized variations in weather conditions within specific, relatively small regions. Unlike broader regional or macro-climate predictions, which provide generalized weather information for large areas, micro-climate predictions focus on understanding the intricacies of weather patterns within smaller, more homogeneous areas. 

Previous works \cite{kumar2021micro, ajani2023greenhouse, eleftheriou2018micro, moonen2012urban} have shown the importance of microclimate prediction for example in \cite{kumar2021micro} an example of a farmer has been discussed, farmer's decision to fertilize their fields hinged on information from a weather station located 50 miles away. This reliance on distant weather data led to a critical issue when localized temperature variations resulted in freezing conditions at night, causing substantial crop damage. This situation underscores the crucial role of accurate micro-climate prediction. Apart from agriculture, microclimate prediction is indispensable in many fields such as forestry, architecture, urban planning, ecology conservation, and maritime activities and it plays a critical role in optimizing decision-making and resource allocations. It empowers stakeholders to make informed decisions and adapt to localized climate variations effectively and thus can play a pivotal role in addressing climate change.


Let's consider predicting climate variables for a new farm or location, where we lack historical data. Traditional methods for microclimate prediction often require extensive data collection, including ground-based sensors or weather stations, which can be costly and limited in coverage. To solve this problem we look at zero-shot micro-climate prediction which refers to the task of predicting fine-grained environmental conditions at specific locations without relying on direct observations or measurements.

In this work, we developed a microclimate prediction deep learning model based on the concept of transfer learning, where knowledge acquired from several locations is used to make predictions in another location with limited or no available data.
Transfer learning enables the application of predictive models trained on existing climate data to estimate microclimate variables such as temperature, humidity, wind speed, and solar radiation in previously unmonitored areas.

In these scenarios, usually larger-scale numerical weather prediction (NWP) models are used which provide predictions for larger areas at high resolution. One such model is called the High-Resolution Rapid Refresh (HRRR) model. Our goal here is to use deep learning and transfer learning to make more accurate forecasts than HRRR model \cite{benjamin2016north} for a previously unmonitored location.

Our model is inspired by previous domain adaptation techniques such as \cite{wilson2020multi, ott2022domain, jin2022domain, xiang2023two, he2023domain} to design the transfer component that transfers the knowledge from the locations with abundant training data to the target location.
\section{Methodology}
In this study, we aim to forecast climate parameters over a time horizon $L_y$ starting from the current time $t$. Our task involves predicting climate parameter values $\mathcal{Y}_t = \{y_{t+1}, y_{t+2}, ..., y_{t+L_y} | y_i \in \mathbb{R}\}$. To achieve this, we take as input a limited preceding window of relevant climate parameters, denoted as $\mathcal{X}_t = \{x_{t-L_x}, x_{t-L_{x}+1}, ..., x_{t} | x_i \in \mathbb{R}^n\}$ where $x_t$ represents the $n$ available climate parameters for input at time $t$, and $y_{t'}$ denotes the target climate parameters of interest at time $t'$. Our predictions are specific to a particular location, referred to as the "target station" ($st_{tar}$), characterized by geographic data encompassing latitude, longitude, and elevation, denoted as $\ell(st_{tar})$. We will use the notations $\mathcal{X}_t(st_i)$, $\mathcal{Y}_t(st_i)$ and $\hat{\mathcal{Y}}_t(st_i)$ to denote relevant past climate parameters, target climate parameter and the forecasts values for station $st_i$ at time $t$. The historical dataset ($\mathcal{H}$) consists of past climate parameter values measured at time intervals preceding $t$ for multiple stations $\mathcal{H} = \{(\mathcal{X}_{t'}, \mathcal{Y}_{t'}) | t' < t\}$, forming the foundation for our predictive modeling. In the zero-shot scenario, historical data within $\mathcal{H}$ does not include any information about the target station ($(\mathcal{X}_t(st_{tar}), \mathcal{Y}_t(st_{tar})) \notin \mathcal{H} \; \forall t$). Consequently, we rely solely on available input data from other sources to forecast climate parameters at $st_{tar}$ and the small immediate preceding window of size $L_x$ for the target station. This method addresses this zero-shot prediction problem by developing and evaluating models capable of accurate climate forecasting at $st_{tar}$ in the absence of historical data specific to that location. 
  
\subsection{Model structure}
\begin{figure}[t]
    \centering
    \includegraphics[width=0.7\textwidth]{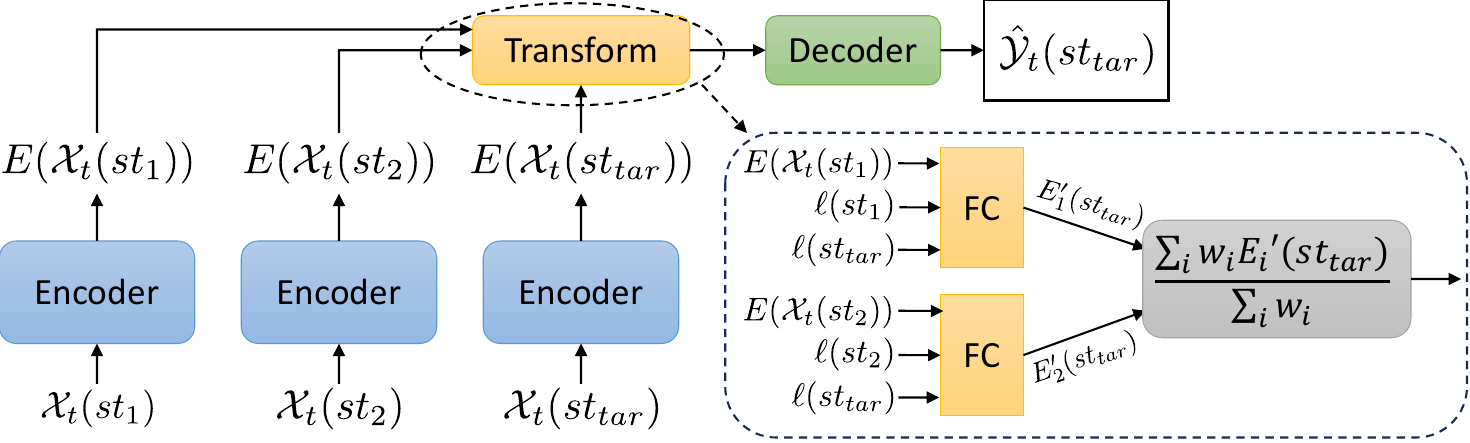}
    \vspace{-2pt}
    \caption{Structure of the transform model. $st_1$ and $st_2$ are the train stations on which the encoder-decoder has been trained, and $st_{tar}$ is the target station which we did not have any training data for. We transform the encoding of $st_1$ and $st_2$ at time $t$ using our Transform component and then pass it through the decoder to get the next 24-hour forecast for the target station.}
    \vspace{-2pt}
    \label{fig:freq_trans_model}
\end{figure}

The model's structure is depicted in Figure~\ref{fig:freq_trans_model}. The central concept behind this architecture is to develop a Transform function capable of extrapolating knowledge from stations for which we possess training data based on their location. This function then transforms this knowledge into the encoding of a target station, even if we lack precise training data for that particular station. Subsequently, we employ the decoder to generate a 24-hour forecast for the target station using this refined embedding. In our implementation, we harnessed the Informer model \cite{informer} for the encoder-decoder, primarily due to its remarkable efficiency in forecasting long sequence time-series data. The Informer model is a highly efficient transformer-based approach for long sequence time-series forecasting that addresses the limitations of traditional Transformer models through innovative mechanisms like ProbSparse self-attention, self-attention distilling, and a generative style decoder, significantly improving time and memory efficiency while maintaining high predictive accuracy.
\subsection{Transform component}
In our model architecture, we utilize a fully connected layer, denoted as $\delta$, which is crucial for transforming the embeddings from source stations to target stations. This transformation leverages information about both source and target locations, in addition to the embeddings of the source stations. The process can be mathematically represented as follows:
$$ E_i'(st_{tar}) = \delta(E(\mathcal{X}_t(st_i)), \ell(st_i), \ell(st_{tar}))$$ 
Here, $E(\mathcal{X}_t(st_i))$ represents the encoder embedding of the source station $st_i$. The output from this layer, $E_i'(st_{tar})$, approximates the encoder embedding for the target station as influenced by $st_i$. Next, we calculate a weighted average of these transformed embeddings from various source stations to get the final estimation of encoder embedding of the target station using the formula:
$$E'(st_{tar}) = \frac{\sum_i w_iE_i'(st_{tar})}{\sum_i w_i}$$
In this equation, the weights $w_i$ as well as the weights in the fully connected layer $\delta$ are dynamically learned during the model's end-to-end training process, along with the rest of the network components.

\subsection{Training procedure}
Our training methodology consists of two distinct phases. Initially, we train the model to forecast the data and we use all train stations. During this phase, the model parameters are updated to capture the global patterns and relationships within the data. Then we pick each of the train stations as the target station, freezing the model parameters except the transform model and the weights of the weighted average combinator and we train these weights.

\section{Results}
In this section, we evaluate the performance of our model in a synthetically generated and real-world dataset. We compare the results against just using the Informer model as encoder-decoder without the use of the Transform component as well as some other baselines explained in Appendix~\ref{sec:app_baselines}.

\subsection{Synthetic datasets}

\begin{table}[t]
\begin{adjustbox}{width=0.8\textwidth,center}
\begin{tabular}{c cc l cc}
\hline\hline
\multirow{2}{*}{\textbf{Model}} & \multicolumn{2}{c}{\textbf{Snthetic Data (MSE$\downarrow$)}} & & \multicolumn{2}{c}{\textbf{AgweatherNet (MSE$\downarrow$)}} \\ 
 &  Full Data & Zero-Shot & & Full Data & Zero-Shot \\
 \cline{2-3}
\cline{5-6}
Last value & $5.2$ & $5.2$ & & $97.66$ & $97.66$\\
Moving average & $38.48$ & $38.48$ & & $63.41$ & $63.41$\\
Persistence model & $10.06$ & $10.06$ & & $26.99$ & $26.99$\\
Auto Regression & $4.95$ & $6.66$ &  & $17.42$ & $17.69$ \\
HRRR & - & - & & $25.53$ & $25.53$\\
Informer & $3.15 \pm 0.04$ & $3.19 \pm 0.03$ & & $17.90 \pm 1.12$ & $19.77 \pm 0.84$ \\
Informer + transform & $\mathbf{2.46  \pm 0.02}$ & $\mathbf{2.60 \pm 0.04}$ & & $\mathbf{14.42 \pm 0.86}$ & $\mathbf{15.02 \pm 0.31}$ \\
\hline\hline
\end{tabular}
\end{adjustbox}
\caption{Average mean squared error results on forecasting synthetic and AgweatherNet data}
\vspace{-12pt}
\label{tab:Results}
\end{table}

To assess the model's ability to recover the original data generation formula, we initiated the process by generating synthetic data. We employed an Ornstein-Uhlenbeck process to simulate this data. This process is explained in appendix Section~\ref{sec:app_syndata}.


In this synthetic data context, our objective was to forecast the next 24 hours of a single variable ($L_y = 24$) based on the last 48 hours of data ($L_x = 48$). We evaluated the performance of each forecasting method by calculating the average mean squared error, both in the complete data case and the zero-shot scenario.

Table~\ref{tab:Results} shows the average mean square error (MSE) results of 10 runs of different models on synthetic data. For the Informer and Informer+transform model, we use the past data of 10 stations as training and we forecast the values for a target station that were not present in the training set. For the Auto Regression models, in the full data scenario, we use the conventional scenario where we give the past data of the target station as training, and in the zero-shot scenario, we train the model using past data of the closest station and use the coefficients for forecasting the target station. For the more basic baseline models (Last value, Moving average, Persistence model) we do not need training data so they are valid for zero-shot scenarios.

We also calculated the errors per variable in the formula for the model by decomposing the results. It is noteworthy that a significant portion of the errors can be attributed to the $\beta$ parameter, which serves as the multiplier for long-term random seasonality. This observation aligns with the inherent difficulty of estimating this parameter accurately, given that our data does not encompass the entire cycle of this seasonality component.
\subsection{Real-world dataset}
\begin{figure}[t]
    \centering
    \includegraphics[width=0.48\textwidth]{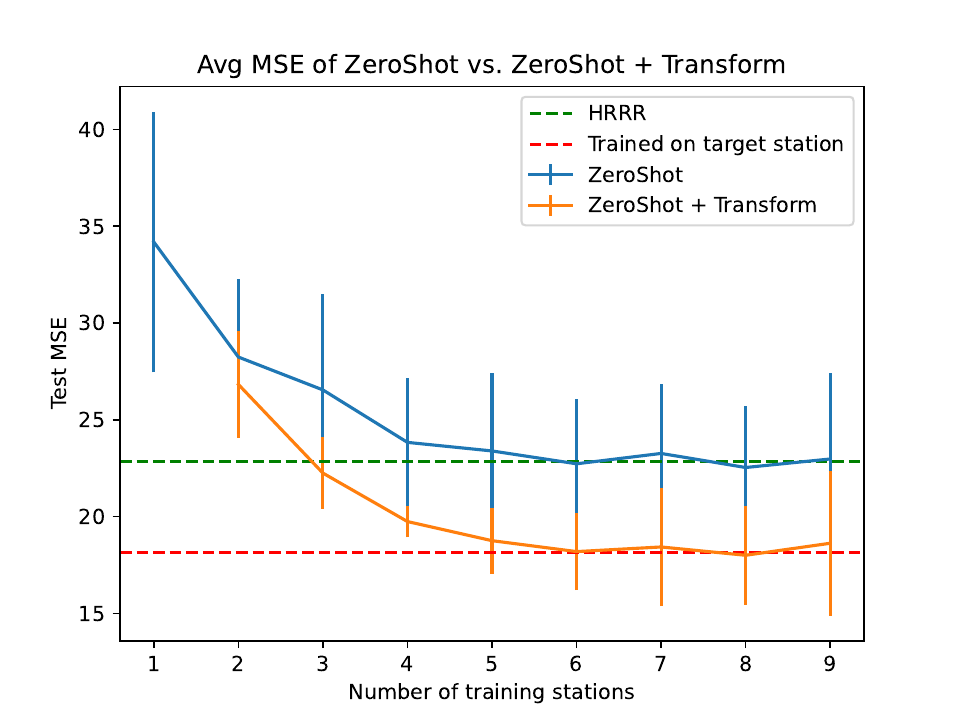}
    \includegraphics[width=0.48\textwidth]{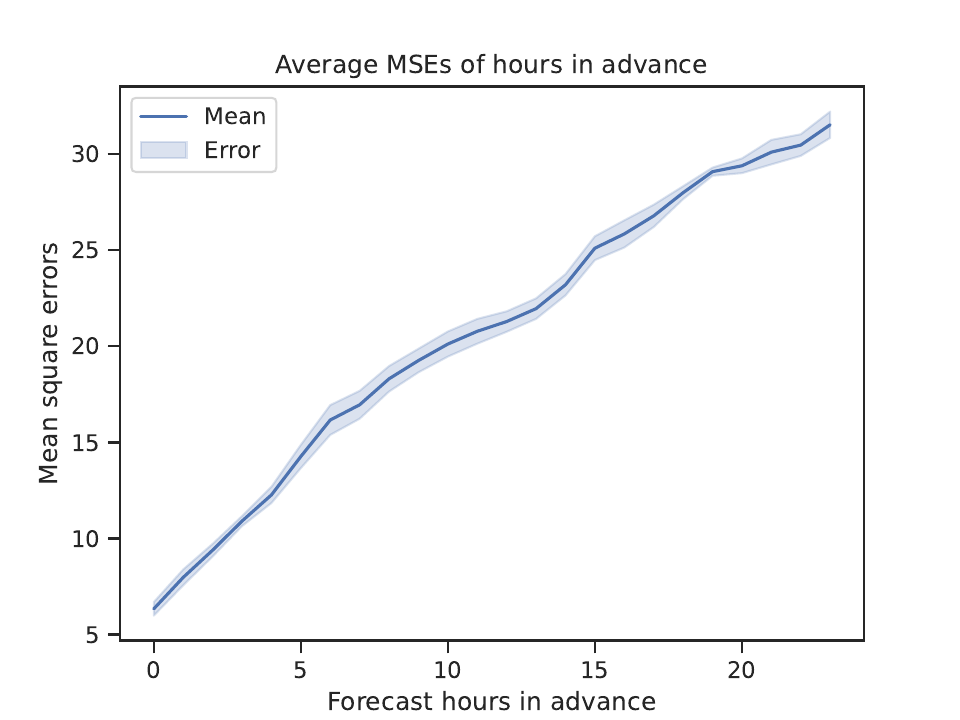}
    \vspace{-2pt}
    \caption{Left: Informer model with and without the transform component when more and more training stations are added. The test mean squared error of HRRR model and when the Informer model is trained and tested on the same station data are shown. The error bars show the standard deviation of 5 runs of the models. Right: the average MSE values across multiple days for each of the 24 forecast hours.}
    \vspace{-2pt}
    \label{fig:stations_transform}
\end{figure}
\begin{figure}[t]
    \centering
    \includegraphics[width=0.45\textwidth]{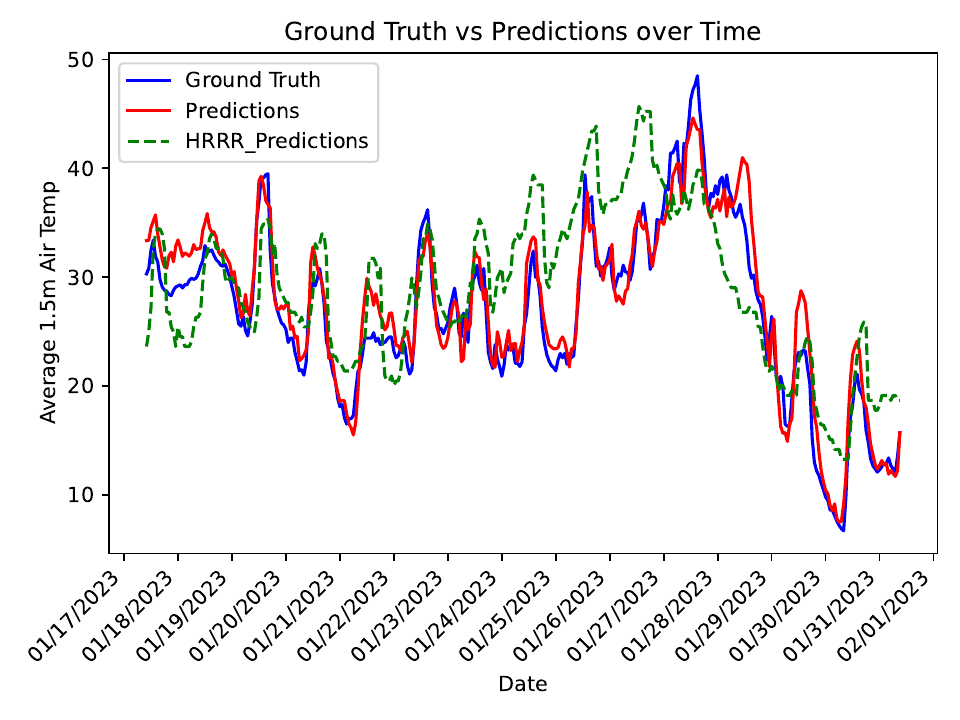}
    \vspace{-2pt}
    \caption{Predictions of our best zero-shot model compared with HRRR predictions and ground truth on the last two weeks of January 2023.}
    \vspace{-2pt}
    \label{fig:PredvsGround}
\end{figure}
To assess the performance of our models, we acquired hourly weather station data from AgWeatherNet\footnote{http://www.weather.wsu.edu} for 10 stations spanning from January 1, 2020, to January 31, 2023. You can find a comprehensive list of these features and the corresponding weather stations in Appendix~\ref{sec:app_agweather}.

Using the data from the past 48 hours ($L_x = 48$), we developed forecasts for the next 24 hours ($L_y = 24$) regarding the Average 1.5-meter air temperature. We then calculated and reported the mean squared error for each hour of the prediction. 

Table~\ref{tab:Results} presents the average Mean Squared Error (MSE) results from five runs of our model, along with comparisons to multiple baseline models. Notably, the Informer model equipped with the transform component, trained on eight stations, demonstrates superior performance, outperforming both the baseline models and the standard Informer model in a zero-shot learning context. The Mean Absolute Error (MAE) results for this dataset are given in Appendix Table~\ref{tab:app_Agweather_mae}.

In Figure~\ref{fig:stations_transform}, we present the average Mean Squared Error (MSE) results for three stations as we progressively include more training data. We compare our model's performance to that of the HRRR model, typically employed in zero-shot scenarios when insufficient training data is available for a specific location. We also contrast this with the conventional scenario where ample data is accessible for both training and testing at the target station. 
Notably, our findings reveal that our model surpasses the HRRR model's performance with just three training stations included. With the inclusion of six training stations, our model performs close to the conventional scenario, where abundant data is available for the target station. This performance can be attributed to the substantial training data available from other stations for our encoder-decoder architecture and because the Transform component can acquire the necessary knowledge to accurately convert the embedding of stations with abundant training data to the embedding of target station, thereby enhancing the accuracy of forecasts for the target station. On the right side of Figure~\ref{fig:stations_transform}, we present the average Mean Squared Error (MSE) for each hourly forecast over 24 hours generated by our model. As anticipated, the accuracy of the forecasts diminishes as we extend the prediction horizon into the future.

In Figure~\ref{fig:PredvsGround}, we showcase the predictions made by our optimal zero-shot model for the final two weeks of January 2023 for the weather station BoydDist, compared with the HRRR model's predictions for the same timeframe. Ground truth values are also presented for comparison. Our model's forecasts align more closely with the ground truth than those of the HRRR model.

\section{Conclusion and future work}
In this paper, we introduced a novel model employing a transformation mechanism designed to extrapolate location embeddings from areas with abundant training data to target locations lacking such data, thus enabling accurate forecasts for previously untrained locations. We demonstrated the model's efficacy in generating precise predictions using both synthetically generated and real-world datasets for previously unmonitored locations, surpassing the performance of the HRRR model in this specific scenario.

In practical applications, when confronted with a new location bereft of weather sensors, we can identify nearby weather stations and leverage our model to make predictions for this unmonitored site. In our future research, we aim to enhance our transformation function by incorporating additional location-specific information and introducing a more intricate structure. We also intend to extend the model's evaluation to a wider array of weather datasets, and more forecasting models. We will also conduct comprehensive analyses, including theoretical proofs, on its performance using synthetically generated datasets.

\bibliographystyle{plain}
\bibliography{references}
\clearpage

\appendix
\section{Synthetic data}
\label{sec:app_syndata}
\begin{table}[t]
    \centering
    \begin{tabular}{|c|c|c|}
        \hline
        \textbf{Parameter} & \textbf{Description} & \textbf{Value/Range} \\\hline
        $\kappa$ & Mean averting parameter of OU & $0.5$\\\hline
        $\sigma(t,x)$ & Noise of OU, we used a constant value & $5$ \\\hline
        $dW(t,x)$ & increment of a normal Brownian motion & $\mathcal{N}(0,1)$\\\hline
        $a(x)$ & Overall mean of OU & $[0 - 20]$ \\\hline
        $b(x)$ & Trend weight & $[0.0 - 0.01]$ \\\hline
        $\omega_1$ & seasonal weather change frequency & $\frac{2\pi}{24*365.25}$\\\hline
        $\omega_2$ & random long-term change frequency & $\frac{0.7\pi}{24*365.25}$\\\hline
        $\omega_3$ & daily weather change frequency & $\frac{2\pi}{24}$\\\hline
        $\theta_1(x)$, $\theta_2(x)$, $\theta_3(x)$ & location-based shift in seasonality & $[0 - \pi]$ \\\hline
        $\alpha(x)$, $\beta(x)$, $\gamma(x)$ & location-based weight of each seasonality & [0 - 15]\\\hline
    \end{tabular}
    \caption{List of parameters, their description and the value or value range we used for them in our experiments}
    \label{tab:app_synparams}
\end{table}
We use an Ornstein-Uhlenbeck process to generate the synthetic data, this is a realistic scenario used before for modeling weather data\cite{mraoua2007temperature, esunge2020weather}. Our formulation is given below:

\begin{align*}
    dT(t,x) &= \big[\frac{d\hat{T}(t,x)}{dt} + \kappa (\hat{T}(t,x) - T(t,x))]dt + \sigma(t,x)dW(t,x)\\
    \hat{T}(t,x) &= a(x) + b(x)t + \alpha(x)sin(\omega_1 t + \theta_1(x)) + \beta(x)sin(\omega_2 t + \theta_2(x)) + \gamma(x)sin(\omega_3 t + \theta_3(x))\\
    \sigma(t,x) &= c
\end{align*}
Which can be generated using the Euler approximation:
$$ T_{i+1}(x) = T_i(x) + \hat{T}_i'(x) + \kappa(\hat{T}_i(x) - T_i(x)) + \sigma_i(x)z_{i,x}$$
In Table~\ref{tab:app_synparams} we give a description of each of these variables and the value/value range used in our experiments.

For the variables that depend on the location such as $a(x)$, $(b(x)$ ... we generate them such that for locations that are close to each other they will have approximately the same values so for $f(x)$ where $f()$ can be any of the parameters above that depend on $x$ we use a linear combination of locations to get values close to each other for similar locations so we have the following formula for generation:
$$ f(x) = \mathcal{N}(\sum_i w_i x_i, \sigma) \times \text{range}_f + \text{min}_f $$
So we generate normally distributed values with standard deviation of $\sigma$ around this limited combination of $x$ values and then scale it to match the range defined in Table~\ref{tab:app_synparams} for this variable.

So given this procedure, we first generate random locations (20 for our experiments) from a uniform distribution, follow the formulas above to generate the parameter values for each location, and then use the OU formulation to generate the final values for that location. For our experiments, we generated hourly data using this procedure for 4 years.
\section{Agweather data}
\label{sec:app_agweather}
\begin{table}[t]
    \centering
    \begin{tabular}{|c|c|c|c|}
        \hline
        \textbf{Station} & \textbf{Elevation} & \textbf{Latitude} & \textbf{Longitude} \\\hline
        BoydDist & 2000 & 47.89 & -120.07 \\\hline
        Harrington & 2170 & 47.39 & -118.29 \\\hline
        PoulsboS & 121 & 47.66 & -122.65 \\\hline
        Seattle & 30 & 47.66 & -122.29 \\\hline
        Addy & 1707 & 48.32 & -117.83 \\\hline
        Almira & 2650 & 47.87 & -118.89 \\\hline
        Broadview & 1492 & 46.97 & -120.5 \\\hline
        Grayland & 21 & 46.79 & -124.08 \\\hline
        Langley & 166 & 48.0 & -122.43 \\\hline
        Azwell & 810 & 47.93 & -119.88 \\\hline
        Mae & 1220 & 47.07 & -119.49 \\\hline
        McKinley & 1081 & 46.01 & -119.92 \\\hline
        MosesLake & 1115 & 47.0 & -119.24 \\\hline
        SmithCyn & 514 & 46.28 & -118.99 \\\hline
    \end{tabular}
    \caption{Stations that are downloaded from AgweatherNet}
    \label{tab:app_Agweather_stations}
\end{table}
\begin{table}[t]
    \centering
    \begin{tabular}{|c|c|}
        \hline
       Min 2m Air Temperature & Average 1.5m Air Temperature \\ \hline
       Max 2m Air Temperature & 1.5m Dew Point\\ \hline
       1.5m Relative Humidity\% & Total Precipitation in inch\\ \hline
       Solar Radiation W/m & 2m Wind Speed mph\\ \hline
       2m Wind Gust mph & 8-inch Soil Temperature\\ \hline
    \end{tabular}
    \caption{List of the parameters that were downloaded from AgweatherNet for each station}
    \label{tab:app_Agweather_params}
\end{table}

\begin{table}[t]
\begin{center}
\begin{tabular}{c cc}
\hline\hline
\multirow{2}{*}{\textbf{Model}} & \multicolumn{2}{c}{\textbf{AgweatherNet (MAE$\downarrow$)}} \\ 
 &  Full Data & Zero-Shot \\
 \cline{2-3}
Last value & $7.42$ & $7.42$\\
Moving average & $6.19$ & $6.19$\\
Persistence model & $3.88$ & $3.88$\\
Auto Regression & $3.07$ & $3.07$ \\
HRRR & $3.86$ & $3.86$\\
Informer & $3.21 \pm 0.15$ & $3.41 \pm 0.24$ \\
Informer + transform & $\mathbf{2.85 \pm 0.05}$ & $\mathbf{2.95 \pm 0.02}$ \\
\hline\hline
\end{tabular}
\caption{Average mean absolute error results on forecasting AgweatherNet data}
\label{tab:app_Agweather_mae}
\end{center}
\end{table}

We downloaded these 10 features listed at Table~\ref{tab:app_Agweather_params} from January 1, 2020, to January 31, 2023 for stations listed in Table~\ref{tab:app_Agweather_stations}.

\section{Baseline models}
\label{sec:app_baselines}
We compare our models against some baseline forecasting models. These models include:\\
\textbf{Last Value}: output the last value seen for the entire forecast window.\\
\textbf{Persistence model}: output the same $F$ previous values for the forecast window.\\
\textbf{Moving average}: predict the average of last window of size $k$ for the future.\\
\textbf{AutoReg}: We train an autoregression\footnote{\url{https://www.statsmodels.org/stable/generated/statsmodels.tsa.ar_model.AutoReg.html}} model that selects models that minimize information criterion (AIC) by selecting trend, seasonal, deterministic, exogenous variables, and lag terms.
\section{Model hyperparameters}
\label{sec:app_hyp}
\begin{table}[t]
    \centering
    \begin{tabular}{|c|c|}
    \hline
        Hyperparameter & Value \\\hline
        batchsize & 32 \\\hline
        inner model embedding size & 2048 \\\hline
        embedding size & 128 \\\hline
        dropout & 0.05 \\\hline
        learning rate & 0.0001 \\\hline
        loss & mse \\\hline
        number of heads & 8 \\\hline
        patience & 10 \\\hline
        encoder layers & 2 \\\hline
        decoder layers & 1 \\\hline
    \end{tabular}
    \caption{Hyperparameters for Informer + transform model}
    \label{tab:app_hyper}
\end{table}
The hyperparameters used for AgweatherNet for informer and informer+transform model are given in Table~\ref{tab:app_hyper}, we select these hyperparameters using 10\% of data that come after the training set and before the test set.
\end{document}